\documentclass[letterpaper]{article} 
\usepackage{aaai25}  
\usepackage{times}  
\usepackage{helvet}  
\usepackage{courier}  
\usepackage[hyphens]{url}  
\usepackage{graphicx} 
\usepackage{natbib}  
\usepackage{caption} 
\usepackage{algorithm}
\usepackage{algorithmic}

\usepackage{algorithm}
\usepackage{algorithmic}
\usepackage{graphicx}
\usepackage{booktabs}
\usepackage{amsmath}
\usepackage{amssymb}
\usepackage{multirow}
\usepackage{subcaption}
\usepackage{pifont}
\usepackage{colortbl}
\usepackage[table,xcdraw]{xcolor}
\usepackage{color}

\usepackage{esint}
\usepackage{newfloat}
\usepackage{listings}
\DeclareCaptionStyle{ruled}{labelfont=normalfont,labelsep=colon,strut=off} 
\floatstyle{ruled}
\newfloat{listing}{tb}{lst}{}
\floatname{listing}{Listing}
\title{Orchestrating the Symphony of Prompt Distribution Learning for Human-Object Interaction Detection}
\author {
    Mingda Jia\textsuperscript{\rm 1},
    Liming Zhao\textsuperscript{\rm 2},
    Ge Li\textsuperscript{\rm 1}\thanks{Corresponding author},
    Yun Zheng\textsuperscript{\rm 2},
}
\affiliations {
    \textsuperscript{\rm 1}Peking University Shenzhen Graduate School\\
    \textsuperscript{\rm 2}Alibaba Group\\
    2201212832@stu.pku.edu.cn, geli@ece.pku.edu.cn, 
    \{lingchen.zlm, zhengyun.zy\}@alibaba-inc.com
}

\usepackage{bibentry}

\begin{document}

\maketitle

\begin{abstract}
Human-object interaction (HOI) detectors with popular query-transformer architecture have achieved promising performance. However, accurately identifying uncommon visual patterns and distinguishing between ambiguous HOIs continue to be difficult for them. We observe that these difficulties may arise from the limited capacity of traditional detector queries in representing diverse intra-category patterns and inter-category dependencies. To address this, we introduce \textbf{Inter}action \textbf{Pro}mpt \textbf{D}istribution Le\textbf{a}rning (InterProDa) approach. InterProDa learns multiple sets of soft prompts and estimates category distributions from various prompts. It then incorporates HOI queries with category distributions, making them capable of representing near-infinite intra-category dynamics and universal cross-category relationships. Our InterProDa detector demonstrates competitive performance on HICO-DET and vcoco benchmarks. Additionally, our method can be integrated into most transformer-based HOI detectors, significantly enhancing their performance with minimal additional parameters.
\end{abstract}
\begin{figure}[t]
  \centering
   \includegraphics[width=1.0\linewidth]{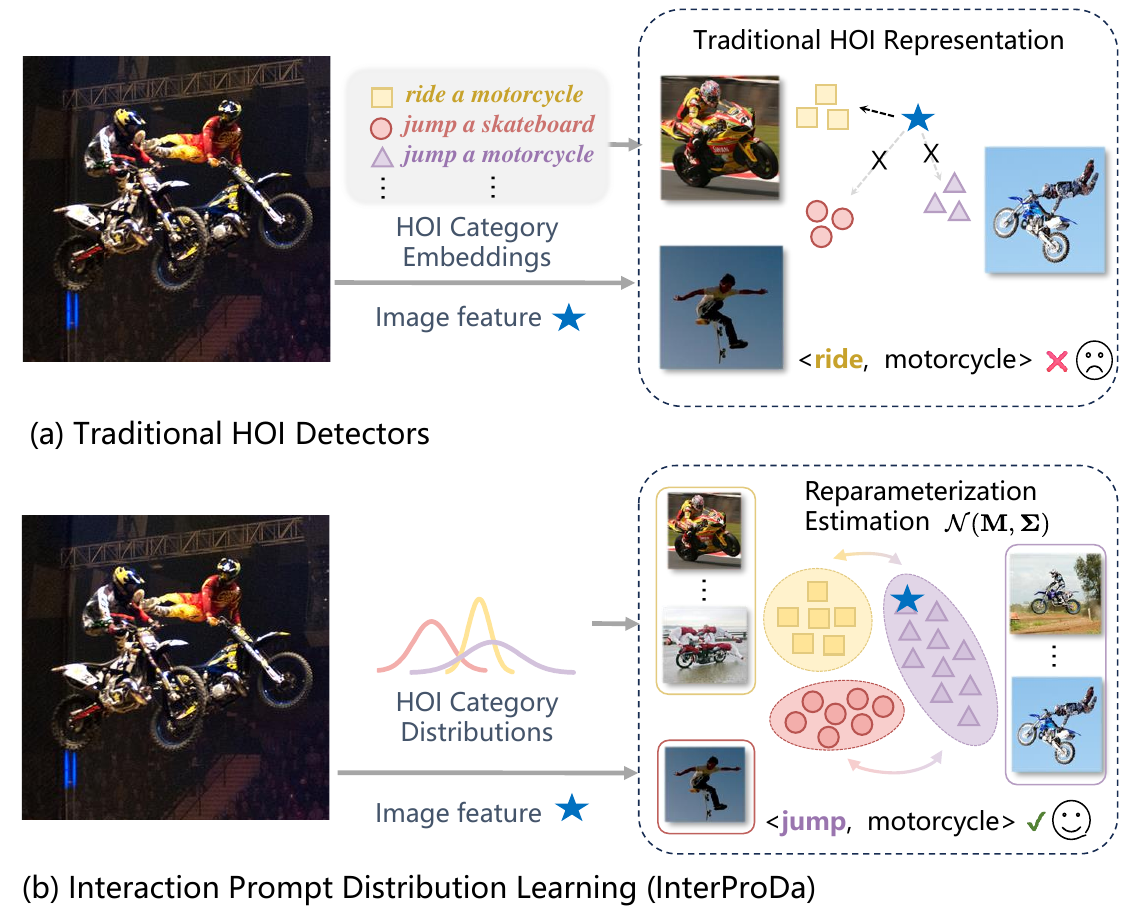}
   \caption{Traditional HOI detectors memorize limited and common visual content, which makes them fragile in recognizing uncommon visual patterns. InterProDa models each HOI category as a distribution to represent unlimited intra-category patterns. We also learn cross-category dependencies with constraints between every distribution. Zoom in for details.}
   \label{fig:figure_intro}
\end{figure} 
\section{Introduction}
Human-object interaction (HOI) detection~\cite{18wacvHOI, kim2021hotr} aims to identify the interactions between humans and objects within a static image and generate predictions in the format $<$ human, interaction, object$>$. Human interactions typically exhibit more complex and diverse visual content than objects. Arguably, the primary difficulty of detecting HOI is understanding the diversity of action patterns. 

Recent advancements in querying-transformer-based HOI methods~\cite{tamura_cvpr2021, ning2023hoiclip, ada_cm, bcom_Wang_2024_CVPR, liao2022gen} have achieved promising performance. Despite this progress, recent HOI detectors still face two difficulties. {Firstly}, they often struggle to recognize interactions involving uncommon visual content. For instance, interactions like $jumping$ a $motorcycle$ suggest a range of visual patterns, such as driving over a ramp or performing mid-air stunts. While humans can easily categorize various patterns, intelligent models tend to misclassify rare visual content into common patterns in the training set, as illustrated in Figure~\ref{fig:figure_intro}. {Secondly}, HOI detectors sometimes confuse visually similar HOIs. 

To address these difficulties and improve zero-shot capabilities, many recent HOI detectors incorporate prior knowledge from pre-trained visual-linguistic models (VLM) into their detection queries by learning discrete~\cite{liao2022gen, ning2023hoiclip} or soft (continuous)~\cite{ada_cm, bcom_Wang_2024_CVPR} prompt embeddings of HOI categories. These category-specific priors shape detector queries to be category queries~\cite{xie2023category}, which learn cross-category dependencies for classification. However, these models still face challenges in capturing diverse intra-category patterns since they do not incorporate any explicit scheme to represent the variations within each query or category.

\begin{figure}[t]
  \centering
   \includegraphics[width=1.0\linewidth]{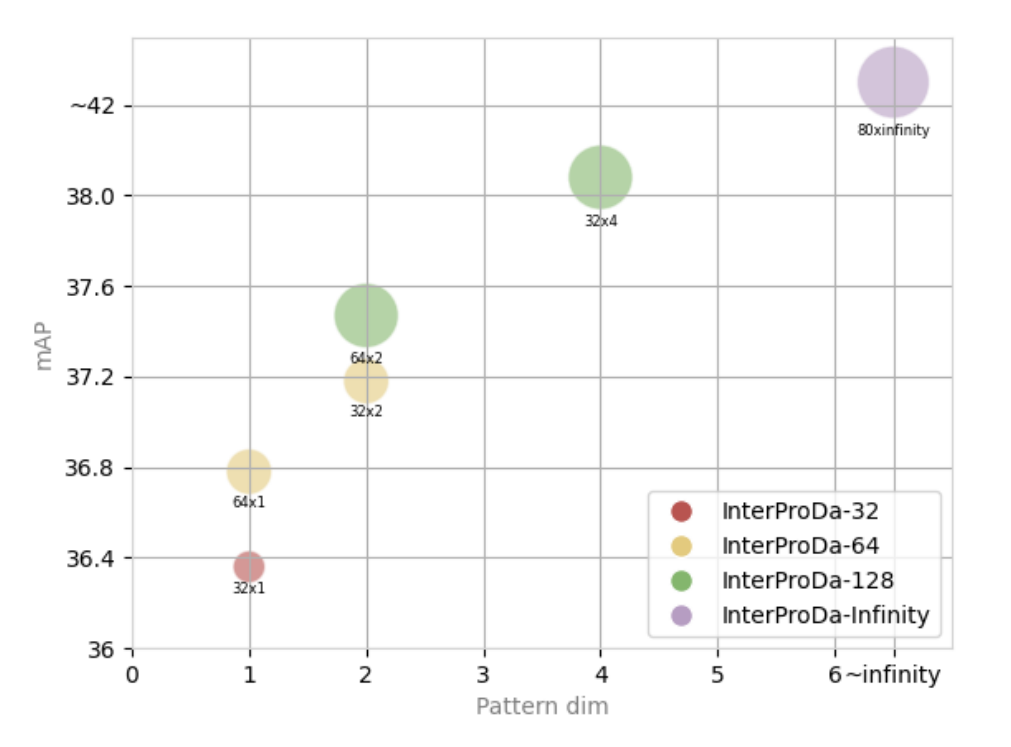}
   \caption{\textbf{Performance comparison w.r.t different settings of decoder queries} on HICO-DET. We expand the query of our model with an additional intra-category pattern dimension. Each model is labeled with a format like $32\times2$, which indicates a model with a category (query) dimension of 32 and pattern dimension of 2. Models with the same color share identical parameters and overall query dimensions (for example, $32\times2$ equals $64\times1$). Comparisons between models of the same color demonstrate that a higher pattern dimension improves performance. More experiment details in Section 4.7.}
   \label{fig:figure_intro_compare_query}
\end{figure}

It motivates us to explore a novel HOI category query structure incorporating an intra-category pattern dimension. While exploring the query pattern design, we observed a noticeable phenomenon that, as shown in Figure~\ref{fig:figure_intro_compare_query}, increasing the query pattern dimension enhances model performance without changing the total number of model parameters. It inspires us to further extend the query pattern dimension, ideally towards infinity, to improve performance even more.

Hence, we propose \textbf{Inter}action \textbf{Pro}mpt \textbf{D}istribution Le\textbf{a}rning (InterProDa), which learns category distribution queries representing approximately infinite intraclass patterns and universal inter-category dependencies in the continuous distribution space. 

In InterProDa, we learn category prompt embeddings as HOI knowledge representation. We first split HOI categories into three groups, including implicit human classes, explicit object classes, and explicit interaction classes. For each group, we design specialized prompt structures. To capture diverse intra-category visual patterns, we learn a collection of soft prompts~\cite{li2021prefixtuningoptimizingcontinuousprompts} for each HOI category. Subsequently, we estimate a unique continuous distribution for each group of prompts. 
A crucial aspect of distribution learning is efficiency. Discriminating between HOI categories involves calculating the marginal likelihood across the multivariate distribution of all categories, which demands substantial computations~\cite {735807}. To optimize this, we assume HOI category representations follow Gaussian distributions, and then we can estimate these distributions efficiently through their means and standard variations.

We capture global inter-category dependencies in the shared feature space of all HOI category distributions. We introduce a novel dynamic orthogonal constraint to enhance space learning with explicit supervision. This constraint increases the distance between ambiguous interaction categories and expands the separation among semantically distinct categories.

In the feed-forward process, we perform sampling on the distribution spaces using a reparameterization trick similar to generative variational auto-encoders (VAE)~\cite{kingma2022autoencodingvariationalbayes} to make the distribution spaces differentiable. We observe that, though the VAE style reparameterization is suitable for generative models, their sampling process with noises makes deterministic models unstable. To mitigate this, we introduce a learnable noise factor during sampling to achieve a balance between learning diverse representations and maintaining stability in HOI detection.

We sample from the learned distribution space and obtain online distribution guidance. By integrating this guidance with learnable queries, we obtain category distribution queries enhancing HOI prediction. InterProDa with our best practice achieves competitive performance on HICO-DET and vcoco. Additionally, our distribution learning pipeline can easily incorporate and boost most HOI detectors with lightweight additional parameters.

\textbf{Contributions.} Our contributions are threefold:
\begin{itemize}
 \item To the best of our knowledge, we are the first to expand query learning to include category distributions, enhancing the recognition of HOIs through uncommon visual content and distinguishing between ambiguous HOIs.

 \item We propose InterProDa, which represents near-infinite intra-category visual patterns and universal inter-category dependencies by category distributions.

 \item InterProDa achieves competitive performance on HOI benchmarks. It can also incorporate and enhance most HOI detectors with minimal additional parameters.
\end{itemize}

\section{Related Works}
\subsection{HOI Detection.} 
HOI detectors with detection transformers have achieved significant success in recognizing human actions toward objects. Both one-stage HOI detectors \cite{tamura_cvpr2021,liao2022gen,ning2023hoiclip, li2023neurallogic} and two-stage detectors~\cite{zhang2022upt, ada_cm, bcom_Wang_2024_CVPR} show their strengths in different detection scenarios. However, recent HOI detectors equipped with traditional decoder queries~\cite{ning2023hoiclip} and category-wise priors~\cite{xie2023category} from VLMs still lack the capacity to represent intra-category dynamics sufficiently. 
 \begin{figure*}[t]
  \centering
   \includegraphics[width=1.0\linewidth]{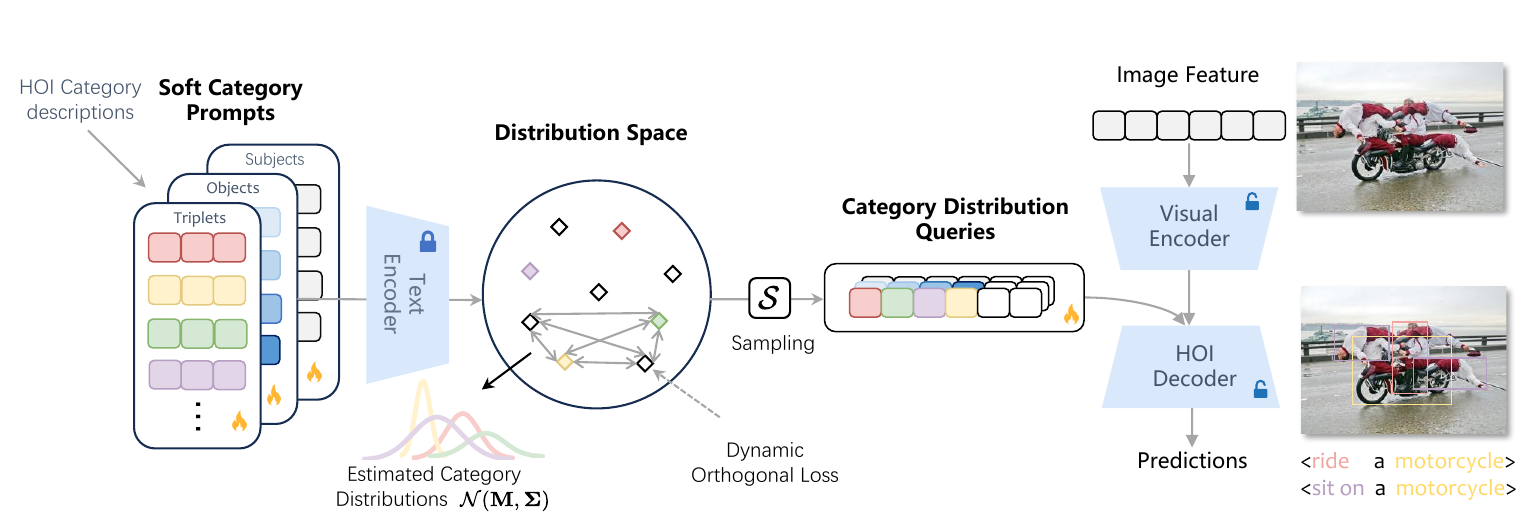}
   \caption{\textbf{The pipeline of InterProDa.} We learn multiple groups of soft prompts for subject, object, and interaction categories. Then, we estimate the distributions of these category prompt embeddings and constrain them in a continuous feature space. Such an approach learns diverse intra-category patterns in each category distribution and captures the universal inter-category dependencies. We sampled from the learned distribution space to obtain a category distribution query to enhance HOI prediction.}
   \label{fig:network}
\end{figure*} 
\subsection{Prompt Learning.}
Prompt learning is an impressive knowledge distillation approach in natural language processing (NLP). It transfers the knowledge of pre-trained models to
downstream tasks by learning proper prompt templates~\cite{Language_models_are_few_shot_learners, jiang-etal-2020-know, shin-etal-2020-autoprompt}. Adopting prompt learning for computer vision, ~\cite{clip, pmlr-v139-jia21b} aligns large-scale visual features with hand-crafted category prompts to train large visual linguistic models (VLM). Leveraging $prompt$ $tuning$, CoOp~\cite{zhou2022coop} learns a soft prompt for visual classification. 

In HOI detection, several works utilise discrete category prompt~\cite{liao2022gen, ning2023hoiclip, Yuan_2023_ICCV}, soft category prompt~\cite{ada_cm, bcom_Wang_2024_CVPR} and tailored multimodal prompts~\cite{mphoi_yang2024open, ting2024CMMP} to improve HOI detection. ADA-CM~\cite{ada_cm} is most related to us, as they tune soft text prompts for HOI detection. Different from them, we learn the distributions of various prompts.
\subsection{Prompt Distribution Learning.}
The varying dynamics of visual content are more complex than language semantics~\cite{he2021maskedautoencodersscalablevision}. To address this, ProDA~\cite{Lu2022Prompt} estimates a category distribution from diverse prompts to achieve generalized adoption of downstream recognition tasks. DreamDistribution~\cite{zhao2023dreamdistributionpromptdistributionlearning} learns a prompt distribution from user input images, allowing the personalized text-to-image (T2I) generation of
novel images. Inspired by their advancement, we leverage prompt distribution to represent theoretically infinite inter-class HOI visual patterns and universal cross-category HOI dependencies. We introduce several novel tailored designs in Section 3 to adopt the prompt distributions to HOI.

\section{Method}
In this section, we term our InterProDa, which can approximately represent infinite intra-category patterns and efficiently capture universal inter-category dependencies. The overall pipeline is illustrated in Figure~\ref{fig:network}.
\subsection{HOI Detection.} 
\label{sec:3.1}
Our HOI detector is shown on the right of Figure~\ref{fig:network}. Given an input image feature map summarized by a visual backbone, a trainable visual encoder will take the feature map and summarize visual content memory ${\mathbf{Z}}$. Then, a set of learnable query embeddings $\boldsymbol{Q}$ will be leveraged by the HOI decoder to ground informative features from ${\mathbf{Z}}$ and generate decoder features $\mathbf{Z}_{dec}$. This process can be expressed by:
\begin{equation}
    \begin{aligned}
          \mathbf{Z}_{dec}= \textrm{Decoder}\left (\textrm{Encoder}({\mathbf{Z}}),\boldsymbol{Q}\right ),
    \end{aligned}
    \label{eq:c_dec} 
\end{equation}
\noindent Several prediction heads will take the decoder features $\mathbf{Z}_{dec}$ and generate HOI predictions ($b_s$, $b_o$, $c_o$, $c_{hoi}$), where $b_s$ is the bounding boxes of subjects in the image, $b_o$ is the object boxes of objects, $c_o$ is the object categories, and $c_{hoi}$ is the categories of HOI triplet combinations.

\subsection{Interaction Prompt Distribution Learning.}
\label{sec:3.2}
\noindent\textbf{Prompt learning.} 
InterProDa aims to learn multiple groups of soft prompts describing the category characteristics for HOI detection. For each HOI category, we learn multiple prompts to represent it. Here, we first present how to learn a single prompt for a category. We feed a randomly initialized natural language prompt to a pre-trained text encoder $\mathcal{F}$ to obtain initial prompt embedding. This initial prompt can be expressed by $\boldsymbol{p}\in\mathbb{R}^{{L}\times{e}}$, where $L$ is the length of tokens in the prompt and $e$ is the embedding dimension. Our best practice is using the openAI CLIP~\cite{clip} text encoder.

Then, the category prompt $\boldsymbol{p}_{c_i}$ of a single category $c_i$ in all the HOI category descriptions $\{c_i\}_{i=1}^{N}$ can be obtained by concatenating the initial prompt $\boldsymbol{p}$ with the CLIP text embeddings of $c_i$, where $N$ is the number of all categories. Specifically, the category prompt $\boldsymbol{p}_{c_i}$ follows the format $[\textrm{PREFIX}]\textbf{V}[\textrm{SUFFIX}]$, with $[\textrm{PREFIX}]$ and $[\textrm{SUFFIX}]$ representing the category-specific prefix and suffix embeddings, respectively. $\textbf{V} \in \mathbb{R}^{M\times{e}}$ signifies the learnable token embedding of the corresponding category, where $M$ represents its token length. The sum of $M$ and the token length of each $c_i$ should be $L$. During prompt learning, $\textbf{V}$ will be updated by the gradient backpropagated through the HOI decoder. 

\noindent\textbf{HOI prompt design.} 
As illustrated in Figure~\ref{fig:network}, we design three HOI-aware structures for category prompts. Given that subjects, objects, and interactions highlight different types of visual content, we employ subject, object, and interaction category prompts to represent these respective concepts.

\textbf{First}, we initialize a set of interaction category prompts  $\mathbf{P}_{int}\in\mathbb{R}^{N_{hoi}\times{L}\times{e}}$, where $\mathbf{P}_{int}$ is the combination of all \{${\boldsymbol{p}_{c}}, c\in C_{hoi}$\}. $C_{hoi}$ represents text descriptions of all HOI combinations in the format of $<$interaction$>$ a/an $<$object$>$. $N_{hoi}$ is the total number of HOI triplet combinations in the training dataset. 

\textbf{Second}, we construct a set of object category prompts $\mathbf{P}_{obj}\in\mathbb{R}^{N_{obj}\times{L}\times{e}}$. Similar to the interaction prompts, $\mathbf{P}_{obj}$ is the collection of all \{${\boldsymbol{p}_{c}}, c\in C_{obj}$\}, $N_{obj}$ is the total number of object categories. Here, we set $C_{obj}$ as the text labels of object categories. 

\textbf{Finally}, for subject prompts, we do not use the text label $human$ as the category description. Humans in the real world inherit fertile semantics like drivers, farmers, athletes, etc. We construct a set of subject category prompts with anonymous categories, leaving all the tokens in these prompts learnable. Thus, subject prompts $\mathbf{P}_{sub}\in\mathbb{R}^{N_{sub}\times{L}\times{e}}$ are filled with $N_{sub}$ randomly initialized soft prompts, where $N_{sub}$ is set to the same as $N_{obj}$.

\begin{figure}[t]
  \centering
   \includegraphics[width=1.0\linewidth]{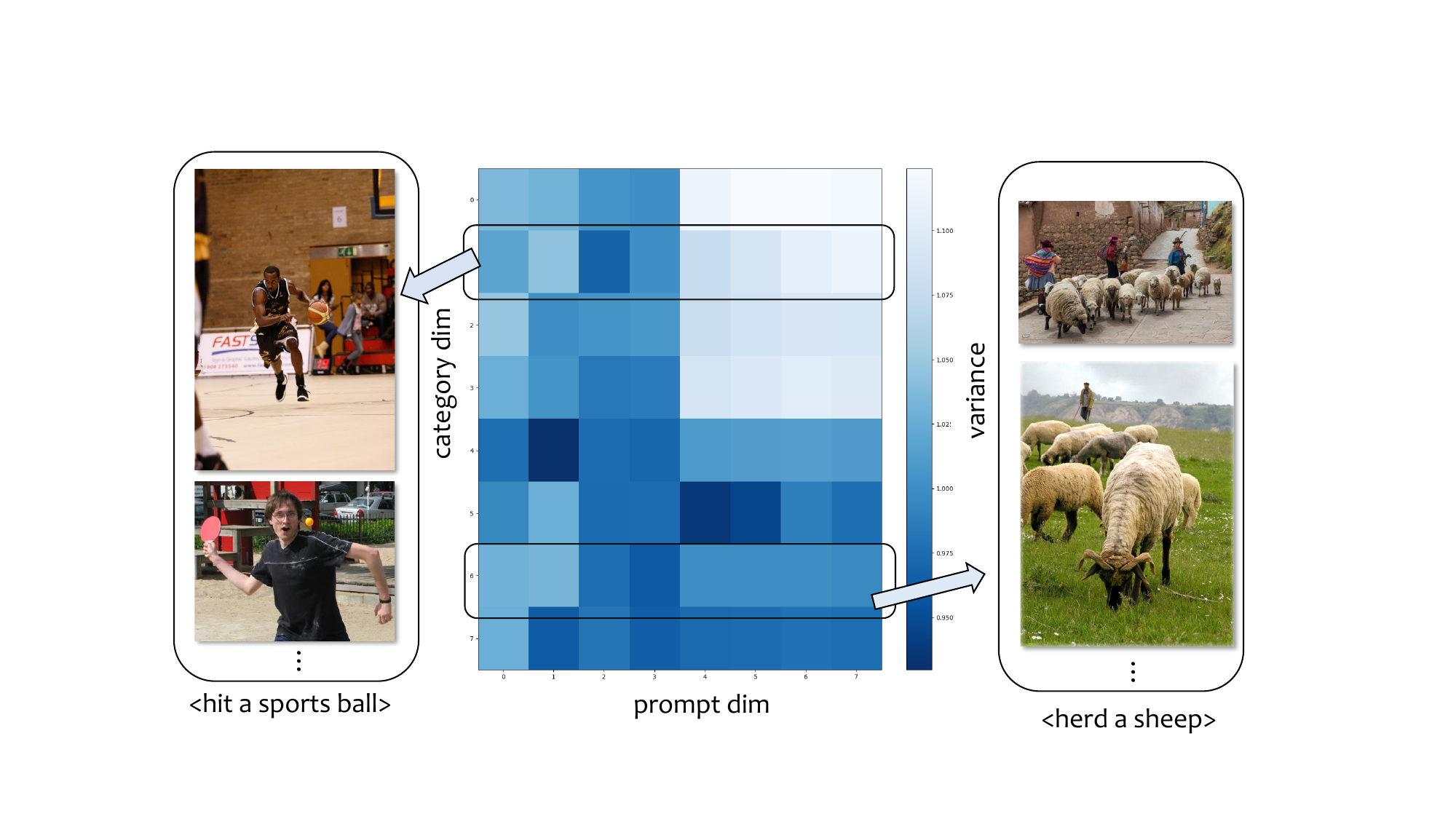}
   \caption{\textbf{Visualization of intra-category variances} of selected prompt distributions learned from HICO-DET. Each row refers to different categories, while each token refers to the variance of a single prompt. We select four distributions with the highest average variance and 4 with the lowest variance, respectively. We sort and list them from top to bottom. We also show HICO-DET images that correspond to two of these category distributions, the distributions with high variance always indicate HOI categories with more diverse visual patterns. }
   \label{fig:tsne_intra}
\end{figure}

\noindent\textbf{Learning prompt distribution.} 
InterProDa further learns the distribution of each HOI category by estimating it from various category prompts. As the CLIP text embeddings with similar semantics remain close in the feature space~\cite{Lu2022Prompt}, our prompt embeddings can be represented by a simple and general distribution. Recent studies found that Gaussian distributions are inherently suitable for feature representations learned by neural networks~\cite{gaussian_union_domain, 9582738, Zhang2022AdversarialRT}. Therefore, we assume that for each HOI category, all the prompt embeddings describing this category follow the same Gaussian distribution. In the following paragraphs, we use the same character to represent the shared process on prompts with different HOI-aware structures.

Specifically, we learn a soft prompt collection consisting of $K$ learnable category prompts for each HOI category, resulting in $\mathcal{P}_{c_i}\in\mathbb{R}^{{K}\times{L}\times{e}}$. Each collection contains $K$ learnable token embeddings $\{\mathbf{V}_k\}_{k=1}^K$. Leveraging the Gaussian distribution prior, we estimate the means $\boldsymbol{\mu}_{c_i} = \mu(\mathcal{P}_{c_i})\in\mathbb{R}^{{L}\times{e}}$ and standard deviations $\boldsymbol{\sigma}_{c_i} = \sigma(\mathcal{P}_{c_i})\in\mathbb{R}^{{L}\times{e}}$ of the normal distribution $\mathcal{N}(\boldsymbol{\mu}_{c_i},\boldsymbol{\sigma}_{c_i}^2)$ for a given category $c_i$. It is worth noticing that $K$ can be theoretically infinite, enabling a distribution estimation that closely approximates the target HOI distribution in the real world, offering the potential ability to encapsulate real-world feature representations. 
\subsection{Distribution Space Learning.}
\label{sec:3.3}
To learn the distributions of all HOI categories simultaneously, we stack all the category distributions $\mathcal{P}_{c_i}$ together to form a universal distribution space $\boldsymbol{\mathcal{P}}\in\mathbb{R}^{N\times{K}\times{L}\times{e}}$. We utilize $\mathcal{N}(\mathbf{M}, \mathbf{\Sigma})$ to represent the multivariate normal distributions $\mathcal{N}(\boldsymbol{\mu}_{c_1:c_N},\boldsymbol{\sigma}_{c_1:c_N}^2)$ describing the distribution space $\boldsymbol{\mathcal{P}}$. This unified space enables our model to effectively capture cross-category dependencies. Building on our design of HOI-aware prompt structures in section 3.2, we independently learn three distribution spaces for subjects, objects, and interactions, represented by $\boldsymbol{\mathcal{P}}_{sub}\in\mathbb{R}^{N_{sub}\times{K}\times{L}\times{e}}$, $\boldsymbol{\mathcal{P}}_{obj}\in\mathbb{R}^{N_{obj}\times{K}\times{L}\times{e}}$ and $\boldsymbol{\mathcal{P}}_{int}\in\mathbb{R}^{N_{hoi}\times{K}\times{L}\times{e}}$.

\noindent\textbf{Space learning constraint.} 
We propose a novel orthogonal constraint with dynamic margins to explicitly guide the learning procedure of the distribution spaces. We aim to maintain that different distributions in the distribution spaces always focus on distinct HOI concepts. In contrast, the distributions of semantically close categories should be closer in the distribution space. The orthogonal constraint calculates a dynamic margin between different distributions according to their cosine similarity:
\begin{equation}
    \begin{aligned}
           \textrm{cos}(i, j) = 
        \dfrac{\left |{\bar{\mathcal{P}}}_{c_i}, {\bar{\mathcal{P}}}_{c_j} \right |} {\|{\bar{\mathcal{P}}}_{c_i}\|_1 \|{\bar{\mathcal{P}}}_{c_j}\|_1 + \epsilon }, i \neq j,
    \end{aligned}
    \label{eq:featureloss}
\end{equation}
where $\bar{\mathcal{P}}$ means the mean-pooled prompt embeddings of each category with shape $\mathbb{R}^{1\times{L}\times{e}}$. Then, their dynamic margin $\Delta$ can be calculated by $\Delta(i, j) = \alpha(1 -\textrm{cos}(i, j)$, where $\alpha$ is a learnable tuning parameter. Finally, the dynamic orthogonal loss can be expressed by:
\begin{equation}
    \begin{aligned}
      \mathcal{L}_{do} = \sum_{i=1}^{N} \sum_{j=i+1}^{N} \left(\max\left(\epsilon, \Delta(i, j) - \textrm{cos}(i, j) \right)\right)^2.
    \end{aligned}
    \label{eq:featureloss}
\end{equation}
As the different semantic concepts of subjects do not vary too much as that of objects or interactions, we only calculate  $\mathcal{L}_{do}$ in $\boldsymbol{\mathcal{P}}_{obj}$ and $\boldsymbol{\mathcal{P}}_{int}$.

\noindent\textbf{Space aggregation.} A significant challenge of HOI distribution space learning is parameter efficiency. To address this issue, we propose a tailor-made distribution aggregation module to compress the original space to a compressed representation $\hat{\boldsymbol{\mathcal{P}}}\in\mathbb{R}^{N'\times K\times L\times e}$, where $N' < N$ represents the number of compressed category distributions. The original $N$ prompt distributions representing explicit HOI categories are summarized into high-level categories aggregating semantically similar categories. 

\noindent\textbf{Sampling from distribution space.}
Each sampling process on the distribution of high-level category $c$ is represented by $\boldsymbol{\mathcal{S}}(\hat{\mathcal{P}_c}) = \hat{\boldsymbol{\mu}}_{c} + \boldsymbol{n}\hat{\boldsymbol{\sigma}}_{c}$, in which $\boldsymbol{n}$ is White Gaussian noise randomly sampled from $\mathcal{N}(0, 1)$. 

To solve the noise in VAE-style reparameterization, we attempt to repeat $\hat{\boldsymbol{\mu}}$ as the estimation of the distribution. However, it will limit the ability to represent fertile HOI patterns. We aim to achieve a balance by adding a learnable scaling factor $\gamma$ on $\boldsymbol{n}$. The sampling process is then modified to $\boldsymbol{\mathcal{S}}(\hat{\mathcal{P}_c}) = \hat{\boldsymbol{\mu}}_{c} + \gamma\boldsymbol{n}\hat{\boldsymbol{\sigma}}_{c}$.

To obtain distribution samples that can sufficiently represent the original distribution, we sample each category distribution for $N_s$ times to obtain the final distribution sampling for each category. This final sampling is expressed by $\hat{\left [{\mathcal{P}}\right ]}_c = \prod_{i = 1}^{N_s} (\hat{\boldsymbol{\mu}}_{c} + \gamma\boldsymbol{n_i}\hat{\boldsymbol{\sigma}}_{c}) $, where $N_s < K$. $\hat{\left [{\mathcal{P}}\right ]}_c$ has shape $\mathbb{R}^{N_s\times L\times e}$. $\prod$ represents concatenating $N_q$ samples of the same distribution along a new dimension. The sampling of the universal HOI category space can be represented by $\hat{\left [\boldsymbol{\mathcal{P}}\right ]}$, with shape $\mathbb{R}^{N\times{N_s}\times L\times e}$. Such sampled distribution space representations are suitable for the feed-forward process and can be updated by propagated gradients.

\subsection{Model Implementation.} 
\label{sec:3.4}
\noindent\textbf{Knowledge distillation.} We implement our HOI decoder with two sequential transformer decoders, including instance decoder and interaction decoder, according to ~\cite{ning2023hoiclip}. The instance decoder and interaction decoder are equipped with learnable instance queries $\boldsymbol{Q}_{ins}\in\mathbb{R}^{(N_{q}\times N_s)\times{C}}$ and interaction queries $\boldsymbol{Q}_{int}\in\mathbb{R}^{(N_{q}\times N_s)\times{C}}$. 

Given the learned category distribution spaces $\boldsymbol{\mathcal{P}}_{sub}$, $\boldsymbol{\mathcal{P}}_{obj}$ and $\boldsymbol{\mathcal{P}}_{int}$, we will sample from them during both training and inference. The sampling generates subject-aware, object-aware, and interaction-aware distribution guidance, termed $\hat{\left [\boldsymbol{\mathcal{P}}\right ]}_{sub}$, $\hat{\left [\boldsymbol{\mathcal{P}}\right ]}_{obj}$ and $\hat{\left [\boldsymbol{\mathcal{P}}\right ]}_{int}$ respectfully. Then, we will take mean pooling on their token dimension and map them to the hidden space of HOI queries by learnable linear layers. We set the compression target $N'$ of our space aggregation module to be the same as $N_q$. Finally, we integrate the distribution guidance and the HOI queries by:

\begin{equation}
    \begin{aligned}
          \boldsymbol{Q}_{d}^{ins} &=   \boldsymbol{Q}_{ins} \oplus \textrm{Linear}(\textrm{cat}(\hat{\left [\boldsymbol{\mathcal{P}}\right ]}_{sub}, \hat{\left [\boldsymbol{\mathcal{P}}\right ]}_{obj})), \\
          \boldsymbol{Q}_{d}^{int} &=   \boldsymbol{Q}_{int} \oplus \textrm{Linear}(\hat{\left [\boldsymbol{\mathcal{P}}\right ]}_{int}),
    \end{aligned}
    \label{eq:c_dec} 
\end{equation}
where $\oplus$ refers to weight addition. Then, the obtained distribution queries $ \boldsymbol{Q}_{d}^{ins}$ and $ \boldsymbol{Q}_{d}^{int}$ will be fed into the corresponding HOI decoders for further process.

\noindent\textbf{Training.}
\label{sec:3.5}
We train InterProDa with an end-to-end fully tuning approach. Both the interaction distribution learning process and the HOI detectors are tuned simultaneously, with a multi-task loss:
\begin{equation}
    \begin{aligned}
          \mathcal{L} = \mathcal{L}_{HOI} +\lambda\mathcal{L}_{do},
    \end{aligned}
    \label{eq:totalloss}
\end{equation}
where $\mathcal{L}_{HOI}$ is the HOI set-prediction loss according to~\cite{tamura_cvpr2021} and $\lambda$ is a hyperparameter. 
\begin{table}[H]
\centering
\scalebox{0.75}{
\begin{tabular}{lccccc}
\hline
\multicolumn{1}{c}{\multirow{2}{*}{Method}} &
  \multicolumn{1}{c}{\multirow{2}{*}{Feature}} &
  \multicolumn{3}{c}{HICO-DET (Default)} &
  \multicolumn{1}{c}{vcoco} \\
  \multicolumn{1}{c}{} &
\multicolumn{1}{c}{Extractor} &
 \multicolumn{1}{c}{\textit{Full}} &
  \multicolumn{1}{c}{\textit{Rare}} &
  \multicolumn{1}{c}{\textit{Nonrare}} &
  \multicolumn{1}{c}{\textit{$AP_{role}^{\#1}$}}\\ \midrule
\specialrule{0em}{1.5pt}{1.5pt}
\midrule
  \multicolumn{1}{l}{HOTR \cite{kim2021hotr}} &
  \multicolumn{1}{c}{R50} &
     25.10 &
  17.34 &
  \multicolumn{1}{c}{27.42} &
  55.2 \\
  \multicolumn{1}{l}{QPIC \cite{tamura_cvpr2021}} &
  \multicolumn{1}{c}{R50} &
     29.07 &
    21.85 &
  \multicolumn{1}{c}{31.23} &
   58.8  \\ 
   \multicolumn{1}{l}{$\textrm{CPC}$ \cite{park2022consistency}} &
  \multicolumn{1}{c}{R50} &
     29.63 &
   23.14 &
  \multicolumn{1}{c}{31.57} &
   63.1  \\ 
   \multicolumn{1}{l}{$\textrm{GENVLKT}$~\cite{liao2022gen}} &
  \multicolumn{1}{c}{R50} &
       33.75 &
   29.25 &
  \multicolumn{1}{c}{35.10} &
    \multicolumn{1}{c}{62.4} \\ 
\multicolumn{1}{l}{PViC \cite{pvic_Zhang_2023_ICCV}} &
  \multicolumn{1}{c}{R50} &
      34.69 
      & 32.14 
      & 35.45 
       & 62.8 
      \\ \multicolumn{1}{l}{$\textrm{HOICLIP}$ \cite{ning2023hoiclip}} &
  \multicolumn{1}{c}{R50 + C} &
      34.69 
      & 31.12 
      & 35.74 
       & 63.5 
      \\
           \multicolumn{1}{l}{$\textrm{CQL}$~\cite{xie2023category}} &
  \multicolumn{1}{c}{R50} &
       35.36&
   32.97&
  \multicolumn{1}{c}{36.07} &
    \multicolumn{1}{c}{66.4} \\
    \multicolumn{1}{l}{RLIPv2 \cite{Yuan_2023_ICCV}} &
   \multicolumn{1}{c}{R50} &
  35.38 &  
  29.61 &  
  37.10  & 65.9  \\  \multicolumn{1}{l}{LOGICHOI\cite{li2023neurallogic}} &
   \multicolumn{1}{c}{R50} &
  35.47 &  
  32.03 &  
  36.22 & 64.4 \\
    \multicolumn{1}{l}{MPHOI\cite{Cao_2023_ICCV}} &
  \multicolumn{1}{c}{R50 + C\&D} &
     {36.50} 
     & 35.48 
     & {36.80} 
     & 66.2 \\
  \multicolumn{1}{l}{RmLR\cite{Cao_2023_ICCV}} &
  \multicolumn{1}{c}{R50} &
     {36.93} 
     & 29.03 
     & {39.29} 
     &63.7 \\
      \multicolumn{1}{l}{ViPLO \cite{park2023viplo}} &
   \multicolumn{1}{c}{R50 + C} 
  & 37.22 
  & 35.45  
  & 37.75   & 62.2 \\
              \multicolumn{1}{l}{SCTC \cite{jiang2024exploringselfcrosstripletcorrelations}} &
  \multicolumn{1}{c}{R50} &
     37.92 
      & 34.78 
      & 38.86 
      & 67.1
      \\ 
 \multicolumn{1}{l}{CMMP \cite{ting2024CMMP}} &
  \multicolumn{1}{c}{R50 + C} &
     38.14 
      & 37.75 
      & 38.25 
      & 64.0 
      \\ 
  \multicolumn{1}{l}{ADA-CM \cite{ada_cm}} &
  \multicolumn{1}{c}{R50 + C} &
     38.40 
      & 37.52 
      & 38.66 
      &58.5 
      \\ 
       \multicolumn{1}{l}{BCOM \cite{bcom_Wang_2024_CVPR}} &
  \multicolumn{1}{c}{R50 + C} &
     39.34 
      & 39.90 
      & 39.17 
       & 65.8 
      \\ 
            \multicolumn{1}{l}{UniHOI \cite{unihoi}} &
  \multicolumn{1}{c}{R50 + B} &
     40.06 
      & 39.91 
      & 40.11 
     & 65.6 
      \\ 
       \rowcolor{blue!5}
\multicolumn{1}{l}{InterProDa (Ours)} &
  \multicolumn{1}{c}{R50 + C} &
    \textbf{42.67} &
   \textbf{45.21} & 
   \multicolumn{1}{c}{\textbf{41.92}} &
   \multicolumn{1}{c}{\textbf{67.6}} \\  \hline 
\end{tabular}}
\caption{\textbf{Performance comparison with state-of-the-art on benchmark HICO-DET and vcoco.} under fully-supervision setting. R50 refers to ResNet50, C refers to CLIP~\cite{clip}, B refers to BLIP2~\cite{li2022blip}, D refers to Stable Diffusion~\cite{rombach2022highresolutionimagesynthesislatent}. Our model achieves competitive performance, especially 45.21 mAP on the $rare$ split.}
  \label{tab:performance_hico}
\end{table}
 \section{Experiments}
\subsection{Datasets \& Metrics}
\label{sec:4.1}
\textbf{Datasets and benchmarks.} We train and evaluate InterProDa on two widely-used HOI Detection datasets, HICO-DET~\cite{18wacvHOI} and vcoco~\cite{gupta2015visual_vcoco}. HICO-DET includes 80 object categories, 117 verb categories and 600 HOI triplet combinations. We split HICO-DET into 38,118 training\&evaluation images and 9,658 test images according to~\cite{tamura_cvpr2021}. vcoco has 5,400 training images and 4,946 validation images, consisting of 81 object categories, 29 verb categories, and 263 HOI triplet combinations. 

\noindent\textbf{Metrics.} We evaluate mean Average Precision (mAP) as our main performance on HICO-DET, following~\cite{18wacvHOI}. Specifically, for an HOI triplet prediction, we first calculate if the IoUs of the human bounding box and object bounding box between groud-truths are larger than 0.5. Then, we check if the predicted categories of interactions and objects are positive. We report both $Default$ and $Known$ $Object$ (KO) split of HICO-DET.  For vcoco, we report scenario 1 AP.
\subsection{Implementations.}
\label{sec:4.2}
Here, we detail the implementation of our best practices. In prompt distribution learning, we set the number of prompts per category, $K$, to 8, with each prompt having a token length $L$ of 77. The reparameterization sampling number $N_s$ is set to 2 per distribution, and the noise scale $\gamma$ for sampling is initialized at 1e-2. The loss weight $\lambda$ for dynamic orthogonal loss is set to 5e-2, and the dynamic margin tuning parameter $\alpha$ is set to 0.5. We use openAI CLIP-ViT-L/32 text encoder for generating prompt embeddings. 

\begin{table}[H]
\centering
\scalebox{0.75}{
\begin{tabular}{cc|ccc}
\hline   
\multicolumn{1}{c}{Method} & \multicolumn{1}{c}{Type} &
 \multicolumn{1}{c}{\textit{Unseen}} &
  \multicolumn{1}{c}{\textit{Seen}} &
  \multicolumn{1}{c}{\textit{Full} } \\ \midrule 
    \multicolumn{1}{l|}{GEN-VLKT\cite{liao2022gen}} &   \multicolumn{1}{c|}{RF-UC} &  
     21.36 
      & 32.91 
      & 30.56
     \\ 
      \multicolumn{1}{l|}{HOICLIP\cite{ning2023hoiclip}} & \multicolumn{1}{c|}{RF-UC} &  
     25.53 
      & 34.85 
      & 32.99 
      \\ 
      \multicolumn{1}{l|}{ADA-CM\cite{ada_cm}} & \multicolumn{1}{c|}{RF-UC} &  
     27.64 
      & 34.35 
      & 33.01 
      \\ 
  \multicolumn{1}{l|}{BCOM\cite{bcom_Wang_2024_CVPR}} & \multicolumn{1}{c|}{RF-UC} &  
     {28.52} 
      & {35.04} 
      & {33.74}  
      \\     \multicolumn{1}{l|}{CMMP\cite{ting2024CMMP}} & \multicolumn{1}{c|}{RF-UC} &  
   {35.98} 
      & {37.42} 
      & {37.13}  
      \\
      \rowcolor{blue!5}
 \multicolumn{1}{l|}{Ours} & \multicolumn{1}{c|}{RF-UC} &  
     \textbf{36.38} 
      & \textbf{40.88} 
      & \textbf{39.58}  
      \\  \midrule
        \multicolumn{1}{l|}{GEN-VLKT\cite{liao2022gen}} &   \multicolumn{1}{c|}{NF-UC} &  
     25.05 
      & 23.38 
      & 23.71
     \\ 
      \multicolumn{1}{l|}{HOICLIP\cite{ning2023hoiclip}} & \multicolumn{1}{c|}{NF-UC} &  
     26.39 
      & 28.10 
      & 27.75 
      \\ 
      \multicolumn{1}{l|}{ADA-CM\cite{ada_cm}} & \multicolumn{1}{c|}{NF-UC} &  
    32.41 
      & 31.13 
      & 31.39 
      \\ 
  \multicolumn{1}{l|}{BCOM\cite{bcom_Wang_2024_CVPR}} & \multicolumn{1}{c|}{NF-UC} &  
    {33.12} 
      & {31.76} 
      & {32.03}  
      \\
       \multicolumn{1}{l|}{CMMP\cite{ting2024CMMP}} & \multicolumn{1}{c|}{NF-UC} &  
    {33.52} 
      & {35.53} 
      & {34.50}  
      \\\rowcolor{blue!5}
 \multicolumn{1}{l|}{Ours} & \multicolumn{1}{c|}{NF-UC} &  
     \textbf{33.64} 
      & \textbf{36.47} 
      & \textbf{35.50}  
      \\
      \bottomrule
\end{tabular}}
\caption{\textbf{Zero-shot HOI Detection}. RF-UC and NF-UC refer to the rare- and non-rare-first unseen combination settings following~\cite{liao2022gen}.}
  \label{table:zero_shot}
\end{table}
\vspace{-1.0em}
\begin{table}[H]
\centering
\scalebox{0.80}{
\begin{tabular}{c|ccc}
\hline   
\multicolumn{1}{c}{Model setting} &
 \multicolumn{1}{c}{\textit{Full}} &
  \multicolumn{1}{c}{\textit{Rare}} &
  \multicolumn{1}{c}{\textit{Non-rare} } \\ \midrule 
  \multicolumn{1}{c|}{base} &
    36.70 
      & 32.58 
      & 37.93
      \\ \midrule
    \multicolumn{1}{c|}{+ prompt query} &  
     37.47 
      & 34.38 
      & 38.40
     \\ 
      \multicolumn{1}{c|}{+ distribution} &  
     38.08 
      & 36.14 
      & 38.65
     \\ 
      \multicolumn{1}{c|}{+ VLM $\mathcal{F}$} & 
     39.56 
      & 40.91 
      & 39.16 
      \\ 
      \multicolumn{1}{c|}{+ $\mathcal{L}_{do}$} & 
     41.85 
      & 43.80 
      & 41.27 
      \\ 
       \multicolumn{1}{c|}{+ Reparam. $\gamma$} & 
     42.19 
      & 43.36 
      & 41.81 
      \\  \rowcolor{blue!5}
 \multicolumn{1}{c|}{+ prompt design} &  
     \textbf{42.67} 
      & \textbf{45.21} 
      & \textbf{41.92}  
      \\  
      \bottomrule
\end{tabular}}
\caption{\textbf{Ablations} on proposed components.}
  \label{table:ablation_components}
\end{table}
\vspace{-1.0em}
\begin{table}[H]
\centering
\scalebox{0.70}{
\begin{tabular}{cc|cccc}
\hline   
\multicolumn{1}{c}{methods} & \multicolumn{1}{c}{Prior} & 
 \multicolumn{1}{c}{\textit{Full}} &
  \multicolumn{1}{c}{\textit{Rare}} &
  \multicolumn{1}{c}{\textit{Non-rare} } &
   \multicolumn{1}{c}{params}
  \\ \midrule 
  \multicolumn{1}{c|}{HOICLIP~\cite{ning2023hoiclip}} &{CLIP-B} &
   34.69 
      & 31.12 
      & 35.74
      & 66M
      \\   \rowcolor{blue!5}
      \multicolumn{1}{c|}{HOICLIP + ours} & {CLIP-B} &
   \textbf{38.04} 
      & \textbf{35.92} 
      & \textbf{38.68}
      & \textbf{68M}
      \\
 \multicolumn{1}{c|}{ADA-CM~\cite{ada_cm}}  & {CLIP-L} &
     {38.40} 
      & {37.52} 
      & {38.66} & 4M
   \\   \rowcolor{blue!5}
    \multicolumn{1}{c|}{ADA-CM + ours}  &  {CLIP-L} &
     \textbf{41.71} 
      & \textbf{42.92} 
      & \textbf{41.34} &  \textbf{5M}
   \\  \bottomrule
\end{tabular}}
\caption{\textbf{Performances of traditional models equipped with our InterProDa}.}
  \label{table:ablation_with_baselines}
\end{table}
We train the distribution space and the HOI decoder concurrently for 60 epochs with an AdamW optimizer on both HICO-DET and vcoco. We set the initial learning rate to 1e-4 and decays after 40 epochs. We set the random seed at 42. For zero-shot training, we follow the setup in ~\cite{ning2023hoiclip}, and we mask the category priors of prompts for unseen categories and initialize them with 8 placeholder tokens,

Our HOI decoder follows an end-to-end transformer encoder-decoder architecture. The encoder and instance decoder have their hidden dimension of 256, while our interaction decoder has the same hidden dimension with the CLIP hidden dimension following~\cite{ning2023hoiclip}. We construct the HOI detector with 6-layer encoders, 3-layer instance decoders and 3-layer interaction decoders. We set query dimension $N_q$ as 80 for both $\boldsymbol{Q}_{d}^{ins}$ and $\boldsymbol{Q}_{d}^{int}$, and their pattern dimension $N_s$ are set to 2, which is also the number of reparameterization samplings. We implement the distribution space aggregation module with a learnable linear layer, which first permutates the distribution space and then maps the category dimension of the space feature to $N' = N_q$.
\begin{figure}[t]
  \centering
   \includegraphics[width=0.85\linewidth]{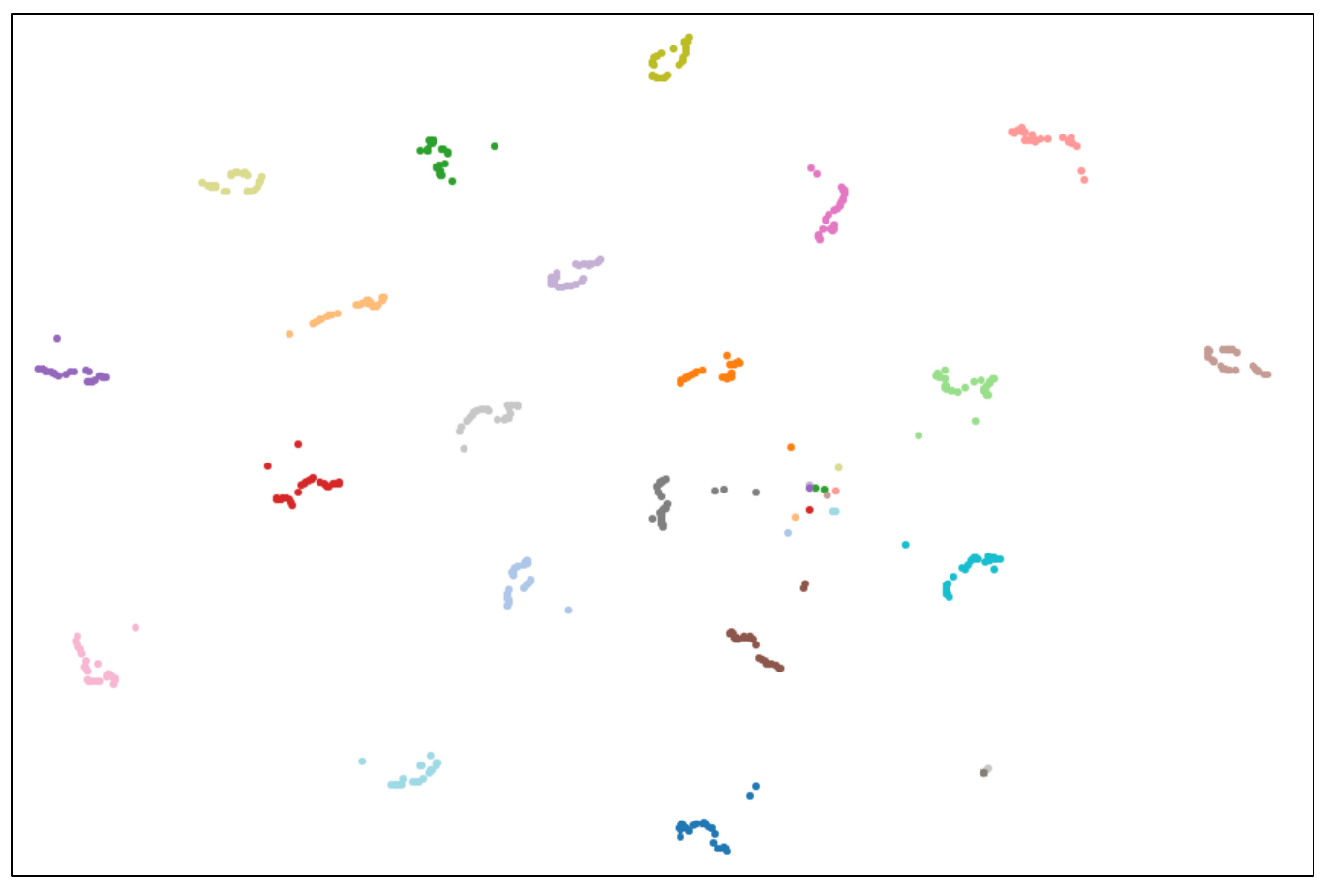}
   \caption{\textbf{T-SNE visualization of learned prompt distributions of 20 random interaction categories on HICO-DET.} Each cluster with the same color refers to the category prompt embeddings belonging to a distinct HOI category distribution. The learned distribution space has clear margins between categories, showing suitable cross-category dependencies. Best viewed in color.}
   \label{fig:tsne}
\end{figure} 
\subsection{Performance comparison}
\label{sec:4.3}
\noindent\textbf{Fully supervised performance.}  We present a performance comparison between InterProDa and SOTA methods trained or fine-tuned with full supervision on HICO-DET and vcoco, detailed in Table~\ref{tab:performance_hico}. InterProda with a ResNet50 backbone and a CLIP feature extractor achieves new SOTA performances of 42.67 $full$ mAP and 45.21 $rare$ mAP on HICO-DET (default). Noticing that our performance enhancement is more significant in the $rare$ split, it highlights the role of interaction distribution learning in exploring hard and uncommon HOI visual content. For vcoco, InterProDa also achieves SOTA 67.6 scenario 1 mAP.

\noindent\textbf{Zero-shot performance.} As CLIP is utilized in InterProDa in the prompt distribution learning, we also compare zero-shot performance with related works. As shown in Table~\ref{table:zero_shot}, InterProDa outperforms recent SOTA zero-shot HOI detectors on both $unseen$ and $seen$ evaluations than recent method considering prompt learning~\cite{ada_cm,bcom_Wang_2024_CVPR}. It is worth noticing that InterProDa with single-modal prompts performs better than CMMP~\cite{ting2024CMMP} with multi-modal prompts, highlighting the effectiveness of prompt distribution learning.

\noindent\textbf{Incorporation with previous works.} We provide portability experimentation on representative works of both one-stage and two-stage HOI detectors, HOICLIP~\cite{ning2023hoiclip} and ADA-CM~\cite{ada_cm}. For HOICLIP, we incorporate InterProDa into its decoder queries. For ADA-CM, we combine its prompt learner with our distribution learning. As illustrated in Table~\ref{table:ablation_with_baselines}, InterProDa enhances HOICLIP~\cite{ning2023hoiclip} to achieve both $full$ and $rare$ mAP with only 2M additional parameters. For ADA-CM, we introduce a significant increase in $rare$ mAP with less than 1M additional trainable parameters.
\begin{table*}[]
\vspace{-1.0em}
\centering
\begin{tabular}{ccc}
\begin{subtable}{.33\linewidth}
\centering
\subcaption{}
\scalebox{0.9}{
\begin{tabular}{c|ccc}
\midrule   
\multicolumn{1}{c}{$\mathcal{L}_{do}$ design} & 
 \multicolumn{1}{c}{\textit{Full}} &
  \multicolumn{1}{c}{\textit{Rare}} &
  \multicolumn{1}{c}{\textit{Non-rare} } \\ \midrule 
    \multicolumn{1}{c|}{w/o $\Delta$} &   
     41.62  
      & 43.75 
      & 40.98 
     \\ 
      \multicolumn{1}{c|}{w / Contra.} & 
     41.60  
      & 43.49 
      & 41.04
      \\  \rowcolor{blue!5}
 \multicolumn{1}{c|}{w/ $\Delta$} &  
     \textbf{41.85} 
      & \textbf{43.80} 
      & \textbf{41.27}
      \\     
      \bottomrule
\end{tabular}}
\end{subtable} 
\hspace{-0.8em}
\begin{subtable}{.33\linewidth}
\centering
\subcaption{}
\scalebox{0.9}{
\begin{tabular}{c|ccc}
\midrule   
\multicolumn{1}{c}{Sampling strategy} & 
 \multicolumn{1}{c}{\textit{Full}} &
  \multicolumn{1}{c}{\textit{Rare}} &
  \multicolumn{1}{c}{\textit{Non-rare} } \\ \midrule 
  \multicolumn{1}{c|}{VAE Reparam.} & 
   42.09  
      & 42.40 
      & 41.99
      \\ \rowcolor{blue!5}
 \multicolumn{1}{c|}{Reparam. w/ $\gamma$}  &
     \textbf{42.67} 
      & \textbf{45.21} 
      & {41.92}
   \\  \rowcolor{white}
   \multicolumn{1}{c|}{Repeat $\boldsymbol{\mu}$}  &
     {42.39} 
      & {43.23} 
      & \textbf{42.14}
   \\  \bottomrule
\end{tabular}}
\end{subtable} 
\begin{subtable}{.33\linewidth}
\centering
\subcaption{}
\scalebox{0.9}{
\begin{tabular}{c|ccc}
\midrule   
\multicolumn{1}{c}{Space basis} & 
 \multicolumn{1}{c}{\textit{Full}} &
  \multicolumn{1}{c}{\textit{Rare}} &
  \multicolumn{1}{c}{\textit{Non-rare} } \\ \midrule \rowcolor{white}
    \multicolumn{1}{c|}{Naive} &
   {40.94} 
      & {40.25} 
      &  {41.31}
      \\ 
      \rowcolor{blue!5}
  \multicolumn{1}{c|}{Gaussian} &
   \textbf{42.01} 
      & \textbf{44.44} 
      &  \textbf{41.29}
      \\ \rowcolor{white}
 \multicolumn{1}{c|}{Fourier}  &
     {42.40} 
      & {44.42} 
      & {41.79}
   \\  \bottomrule
\end{tabular}}
\end{subtable} \\
\begin{subtable}{.85\linewidth}
\centering
\subcaption{}
\scalebox{0.85}{
\begin{tabular}{cccc| cccc| cccc}
\hline   
 \multicolumn{1}{c}{\textit{$K$}} &
 \multicolumn{1}{c}{\textit{Full}} &
  \multicolumn{1}{c}{\textit{Rare}} &
  \multicolumn{1}{c}{\textit{Non-rare} } &
   \multicolumn{1}{c}{\textit{$N_s$}} &
    \multicolumn{1}{c}{\textit{Full}} &
  \multicolumn{1}{c}{\textit{Rare}} &
  \multicolumn{1}{c}{\textit{Non-rare} } &
   \multicolumn{1}{c}{\textit{$\lambda$} } &
     \multicolumn{1}{c}{\textit{Full}} &
  \multicolumn{1}{c}{\textit{Rare}} &
  \multicolumn{1}{c}{\textit{Non-rare} }
   \\  \midrule
  \multicolumn{1}{c}{2} &
    41.62 
      & 43.75 
      & 40.96 
      &  \multicolumn{1}{c}{8} 
      &
    39.79 
      & 40.42 
      & 39.80
      &  \multicolumn{1}{c}{1e-1}  &
    41.62 
      & 43.75 
      & 40.98 
      \\ 
 \multicolumn{1}{c}{4} & 41.83 
      & 43.96 
      & \ \textbf{41.19} 
      &  \multicolumn{1}{c}{4}  
      &
    41.49 
      & 43.71 
      & 40.82
      & \multicolumn{1}{c}{1e-2} &
    41.72 
      & 43.52 
      & 41.19
      \\ \rowcolor{blue!5}
     \multicolumn{1}{c}{8} &
     \textbf{41.93} 
      &  \textbf{44.77} 
      & 41.08 %
      &  \multicolumn{1}{c}{2}   &
     \textbf{41.62} 
      & \textbf{43.75} 
      & \textbf{40.96} %
      &  \multicolumn{1}{c}{5e-2} &
     \textbf{42.09} 
      & \textbf{44.19} 
      & \textbf{41.46} 
      \\ 
      \bottomrule
\end{tabular}}
\end{subtable}
\hspace{-0.8em}
\end{tabular}
\caption{\textbf{Ablations} of (a) designs of dynamic orthogonal loss, (b) sampling strategy, (c) feature space bases and (d) model hyperparameters. All experiments are conducted with ResNet50 backbone and CLIP-L on HICO-DET.}
\label{table:ablation_others}
\end{table*}

\subsection{Component ablation.} 
We evaluate the effectiveness of our proposed components in Table~\ref{table:ablation_components}. The base item refers to the fundamental HOI detector detailed in the implementation details. In the next row, we initialize three sets of learnable prompts without HOI category priors. For the $distribution$ item, we expand these prompts to prompt collections and model them as Gaussian distributions. By mapping the prompts to VLM text embeddings, the model gains higher $rare$ mAP. Equipping the distribution space with $\mathcal{L}_{do}$, the capability to represent fertile visual patterns is simulated. By updating our distribution prompts to HOI-aware prompts with different tailored structures, our model receives more improvement.

\subsection{Analysis on Learned Distributions.}
\label{sec:4.4}
\noindent\textbf{Visualization of intra-category patterns.}
We provide visualizations to analyze the learned intra-category patterns in each HOI category distribution, as shown in Figure~\ref{fig:tsne_intra}. The plots display the standard variances of prompts within each category.  We observe that distributions with higher standard variances tend to correspond to HOI categories with more complex and diverse patterns. For example, in the figure, the category $hit$ a $ball$ covers various activities, such as playing basketball and hitting a ping-pong ball. In contrast, the $herd$ a $sheep$ category show similar poses and spatial arrangements. These distribution characteristics align well with our expectations.

\noindent\textbf{Visualization of inter-category dependencies.}
We also visualize the learned inter-category dependencies in our HOI distribution space, as presented in Figure~\ref{fig:tsne}. The distributions maintain a relatively clear margin between others, making the space representation suitable for HOI determination.

\subsection{Other Ablations.}
\label{sec:4.6}
\noindent\textbf{Dynamic orthogonal constraint.} We compare different designs of $\mathcal{L}_{do}$ in Table~\ref{table:ablation_others} (a). $\textrm{Contra.}$ refers to adding hard contrastive margins without the tuning parameter $\alpha$. It shows the effectiveness of the dynamic margin $\Delta$ in capturing suitable distribution space dependencies.

\noindent\textbf{Different sampling Strategies.} We also test different sampling strategies in Table~\ref{table:ablation_others} (b). VAE Reparam. setting follows~\cite{kingma2022autoencodingvariationalbayes}, and Reparam. w/$\gamma$ indicates our proposed approach. Repeat $\boldsymbol{\mu}$ refers to filling each distribution sample with the mean of the distribution. From the table, our learnable noise factor minimizes the impact of the noise on deterministic inference.

\noindent\textbf{Model hyper-parameters.} We detailed our hyperparameter experiments in Table~\ref{table:ablation_others} (d). For each experiment, we set irrelevant $K$, $N_s$ and $\lambda$ to 2, 2 and 1e-2 by default. It is noticeable that $K$ can be theoretically set to $\infty$ with ideal memory storage, we set $K$ to 8 for model efficiency. 

\noindent\textbf{Different space bases.} We evaluate different space bases in modeling the continuous HOI category representation in Table~\ref{table:ablation_others} (c). $Naive$ means sampling prompts from collections of the category prompt embeddings by max-pooling in each prompt group. $Fourier$ is stated in section 4.8.

\subsection{Exploration on query pattern dimension.}
For the performances illustrated in Figure~\ref{fig:figure_intro_compare_query}, we learn randomly initialized prompt embeddings with different category dimensions and pattern dimensions, like $\mathcal{P}\in\mathbb{R}^{{N_q}\times{N_p}\times{C}}$, where $N_q$ is the category dimension, $N_p$ is the pattern dimension. Orthogonal loss is added to the $N_q$ dimension to distinguish different categories. In the HOI detector, we reshape their queries to $\mathcal{P}\in\mathbb{R}^{({N_q}\times{N_p})\times{C}}$ for calculating $\mathcal{L}_{hoi}$. In this setting, for instance, models with labels ($32\times2$) and ($64\times1$) in Figure~\ref{fig:figure_intro_compare_query} share the same parameter amounts but with different learning processes before being fed into the detector.

\subsection{Prompts in frequency space.}
In the view of signal processing, modeling category embeddings as distributions is actually representing them with Gaussian bases. The orthogonal Fourier basis is another signal basis suitable for reconstructing varying signals due to its continuous nature. We attempt to utilize Fourier transformation to project pooled category prompt embeddings to the frequency domain. This approach also obtains similar performance with using Gaussian distributions.

\section{Conlusion}
We first introduce a novel interaction distribution learning approach, InterProDa, to HOI detection, which learns prompt distributions for HOI categories. InterProDa learns category distribution queries to represent nearly infinite intra-category HOI patterns. It also learns universal cross-category dependencies. InterProDa, to our best practice, achieves new State-of-the-Art performance on HOI detection benchmarks and can significantly boost existing query-based HOI detectors with lightweight additional parameters.

\clearpage
\bibliography{aaai25}

\end{document}